\documentclass[lettersize,journal]{IEEEtran}
\usepackage{amsmath,amsfonts}
\usepackage{algorithmic}
\usepackage{array}
\usepackage[caption=false,font=normalsize,labelfont=sf,textfont=sf]{subfig}
\usepackage{textcomp}
\usepackage{stfloats}
\usepackage{url}
\usepackage{verbatim}
\usepackage{graphicx}
\usepackage{tabularx}
\usepackage{multirow}
\usepackage[colorlinks, linkcolor=blue, citecolor=blue, urlcolor=blue]{hyperref}
\def\BibTeX{{\rm B\kern-.05em{\sc i\kern-.025em b}\kern-.08em
    T\kern-.1667em\lower.7ex\hbox{E}\kern-.125emX}}
\usepackage{balance}
\begin{document}
\title{Modeling of Core Loss Based on Machine Learning and Deep Learning}
\author{Junqi He, Yifeng Wei, Daiguang Jin
\thanks{
The first two authors contributed equally to this work. \textit{Corresponding author: Junqi He}.

The authors are with College of Sciences, China Jiliang University, Hangzhou 310018, PR China (e-mail: hejunqi@cjlu.edu.cn).}}

\markboth{}%
{How to Use the IEEEtran \LaTeX \ Templates}

\maketitle

\begin{abstract}
This article proposes a Mix Neural Network (MNN) based on CNN-FCNN for predicting magnetic loss of different materials. In traditional magnetic core loss models, empirical equations usually need to be regressed under the same external conditions. When the magnetic core material is different, it needs to be classified and discussed. If external factors increase, multiple models need to be proposed for classification and discussion, making the modeling process extremely cumbersome. And traditional empirical equations still has the problem of low accuracy, although various correction equations have been introduced later, the accuracy has always been unsatisfactory. By introducing machine learning and deep learning, it is possible to simultaneously solve prediction problems with low accuracy of empirical equations and complex conditions. Based on the MagNet database, through the training of the newly proposed MNN, it is found that a single model is sufficient to make predictions for at least four different materials under varying temperatures, frequencies, and waveforms, with accuracy far exceeding that of traditional models. At the same time, we also used three other machine learning and deep learning models (Random Forest, XGBoost, MLP-LSTM) for training, all of which had much higher accuracy than traditional models. On the basis of the predicted results, a hybrid model combining MNN and XGBoost was proposed, which predicted through weighting and found that the accuracy could continue to improve. This provides a solution for modeling magnetic core loss under different materials and operating modes.
\end{abstract}

\begin{IEEEkeywords}
core loss, machine learning, deep learning, neural network, Steinmetz Equation.
\end{IEEEkeywords}

\section{Introduction}
\IEEEPARstart{M}{agnetic} components are an important research field in power electronics technology. As an indispensable component in electronic components, they directly affect important performance indicators such as volume, weight, and loss\cite{7570571,278656,9477291,4802625,10700105}. With the continuous development of power electronics technology towards high frequency, high power density, and high reliability, the performance requirements for magnetic components are also increasing. Core loss is an inevitable phenomenon in magnetic components, which can lead to electronic components heating and performance degradation. So accurately predicting the core loss of magnetic components plays an important role in designing electronic components\cite{7395385,10707354}.

The magnitude of magnetic core loss is mainly related to many factors such as operating frequency, operating temperature, magnetic flux density, excitation waveform, magnetic materials, etc., presenting complex nonlinear and linear relationships. At present, the prediction model of magnetic core loss mainly relies on empirical equations such as Steinmetz Equation (SE)\cite{5570437}. The expression for the SE is
\begin{equation}
P = k \cdot f^a \cdot B_{\text{max}}^b
\end{equation}
among them, $P$ is the magnetic core loss per unit volume, $f$ is the frequency of alternating current, $B_{max}$ is the peak value of magnetic flux density, and $k$, $a$, $b$ are the coefficients of experimental data fitting, respectively. The advantage of SE lies in its simplicity and convenience, but its limitations are also significant. SE can only predict when the excitation waveform is a sine wave, and the fitting coefficients may not be the same under different magnetic core materials and operating modes.

In order to predict non sinusoidal magnetic core loss, the improved Generalized Steinmetz Equation (iGSE) was proposed\cite{1196712}, expressed as
\begin{equation}
P = \frac{k_i (\Delta B)^{b - a}}{T} \int_{0}^{T} \left| \frac{d B}{d t} \right|^{a} d t
\end{equation}
\begin{equation}
k_i = \frac{k}{(2\pi)^{a - 1} \int_{0}^{2\pi} |\cos \theta|^{a} 2^{b - a} d\theta}
\end{equation}
where $\Delta B$ is the peak to peak value of magnetic flux density within one cycle, and $k$, $a$, $b$ is the coefficient of SE at the equivalent frequency. iGSE is applicable to magnetic core loss under any waveform, and it corrects the traditional Steinmetz equation by considering the rate of change in magnetic flux density (i.e. $\frac{dB(t)}{dt}$) and $\Delta B$, without relying solely on maximum magnetic flux density and frequency. In addition, there are  Modified SE (MSE), improved-iGSE ({i}$^{2}$GSE), etc\cite{955931,5955126,936396,548774}. They all predict core loss under more complex operating modes by introducing more factors. Although these modified equations compensate for many shortcomings of SE and provide better predictive ability under different conditions, they still cannot fully predict the magnetic core loss under all different conditions. For example, the magnetic core loss under different materials still need to be discussed separately, and the influence of temperature on magnetic core loss has not been taken into account. So the limitations of traditional models have prompted us to seek a new modeling method to improve the accuracy and applicability of magnetic core loss prediction models.

In this context, modeling and predicting magnetic core loss through machine learning has become a very good choice. In recent years, many machine learning models have been applied to predict magnetic core loss\cite{5367762}. However, for traditional machine learning models, there is no good method to regress the sequence of excitation waveforms, and manual feature extraction is still needed to train the model. To address this issue, deep learning models have been introduced. Due to the powerful analytical ability of neural networks for sequences, they have gradually been widely used in predicting magnetic core loss\cite{10608787,10615222,10315145,10567902,9907097,sapkota2024deep}. The main neural network architectures currently proposed for predicting magnetic core loss include CNN, MLP, LSTM, etc\cite{10567902,9907097,10232863}. The newly proposed architectures include KANN\cite{10315145}, PI-MFF-CN\cite{10535744}, etc. The research results demonstrate the enormous potential of neural network-based modeling for predicting magnetic core loss, but also expose some issues: increasing the complexity of the model in order to obtain better prediction results requires more parameters and computational resources for training neural network models, and there may also be overfitting problems; At present, all the proposed models are still unable to predict the core loss of multiple materials under the condition of only training a single model.

In order to solve the above two problems, this paper proposes an MNN architecture based on CNN-FCNN. Compared with existing models, this architecture has a simpler structure, excellent generalization ability, and solves the problem of predicting magnetic core loss of different materials. In addition, we also tried other machine learning models, and ultimately found that both MNN and XGBoost can achieve high-precision prediction of magnetic core loss for different materials. The results indicate that machine learning and deep learning can predict the magnetic core loss of multiple materials while training only one model. Based on data-driven approach, it is possible to establish a high-precision and universally applicable magnetic core loss model for various operating modes.

The rest of this article is organized as follows. Section \ref{sec:2} introduces the methods of dataset pre-processing and the newly proposed neural network model framework, and provides evaluation indicators for the model. Section \ref{sec:3} first presents the prediction results of traditional empirical equations, and then introduces the training methods and prediction results of different machine learning and deep learning models. It can be found that without considering material types, machine learning and deep learning models can achieve smaller errors than empirical equations considering material types. Through analysis, we also found that a hybrid method can be used to obtain better prediction results based on the newly proposed model. Finally, Section \ref{sec:4} provides a summary of this article.

\section{Methodology}
\label{sec:2}

\subsection{Dataset pre-processing}
To compare the gap between machine learning models and traditional models, we trained them using the MagNet\cite{9773372} dataset. In the open-source dataset, we only selected some features for training, as shown in Table \ref{tab1}. The following training results were conducted under this premise. A total of 12400 samples were selected and divided into training set, validation set, and testing set according to a ratio of 7:1.5:1.5. The deep learning model is implemented through pytorch.

\begin{table}
\centering
\caption{Feature selection for training set}
\label{tab1}
\renewcommand{\arraystretch}{1.2} 
\begin{tabular}{ccc}
\hline
Feature                  & Value                     & Data category \\ \hline
Material                 & 3C94, 77, N27, N87            & Nonlinear     \\
Waveform                 & Sine, triangular, trapezoidal & Nonlinear     \\
Temperature(°C)          & 25, 50, 70, 90                & Linear        \\
Frequency(Hz)            & 50000-500000Hz                & Linear        \\
Magnetic flux density(T) & 1x1024 vector                 & Sequence      \\ \hline
\end{tabular}
\end{table}

In deep learning, it is necessary to standardize the features. The standardized formula is
\begin{equation}
z=\frac{x-\mu }{\sigma }
\end{equation}
Where $\mu$ is the mean of the feature and $\sigma$ is the standard deviation of the feature. Standardizing the dataset can help improve the training performance of deep learning models, as the optimization algorithm of the model is quite sensitive to feature scales.

\subsection{Machine learning models}
Random Forest\cite{breiman2001random} and eXtreme Gradient Boosting\cite{10.1145/2939672.2939785} (XGBoost) are two commonly used decision tree based machine learning models that have stable and good performance when dealing with large-scale complex datasets. Random Forest has good resistance to overfitting, with few model parameters and relatively simple; XGBoost has better accuracy, but the model has more parameters and is relatively complex.

Random Forest is an ensemble learning method that constructs multiple decision trees and combines their predictions to improve accuracy and stability. The main idea behind Random Forest is to introduce randomness to reduce model variance, thus enhancing predictive robustness. In the regression task, the calculation steps of a Random Forest are as follows: first, randomly sample subsets from the dataset to build multiple different subsets. Then, for each subset, generate a decision tree. At each split, randomly select a subset of features and use MSE to determine the best split. Finally, take the average of the predicted results as the output.

XGBoost is an algorithm based on boosting trees and the gradient boosting framework. Unlike Random Forest, XGBoost optimizes the model by gradually fitting residuals and utilizing gradient descent. Regularization terms were also introduced to prevent the model from becoming too complex and overfitting. The steps of XGBoost model regression are to initialize the model first, set the initial predicted value as the mean of the target value, train a new tree through the residuals of the current model, then use first-order and second-order degrees to optimize node splitting, and finally update the model with the predicted results of the new tree. Repeat the above steps until the model converges.

Both models require manual feature extraction of the sequence before regression prediction. No matter which machine learning model is chosen, feature engineering is crucial for the effectiveness of pure machine learning methods.

\subsection{MLP-LSTM architecture}
Multi-Layer Perception (MLP) and Long Short-Term Memory (LSTM) are two neural network architectures. MLP can extract features from discrete data through linear transformations and nonlinear mappings. LSTM can be used to process data related to time series, where the input sequence has temporal characteristics and the same time steps, and the waveform sequence of magnetic flux density precisely possesses such characteristics. In recent research, both MLP\cite{9907097} and LSTM\cite{10232863} have been used separately to predict core loss, so using the MLP-LSTM model can combine the advantages of both to make more accurate predictions on the dataset\cite{10567902}.

MLP-LSTM consists of MLP, LSTM, and fully connected layers, and its architecture is shown in the Figure \ref{MLP-LSTM}. MLP extracts features of nonlinear data. LSTM is used to process sequential data. Then concatenate the features and enter the fully connected layer for prediction. The principles of MLP and LSTM models are as follows:

\begin{figure*}
\centering
\includegraphics[width=\textwidth]{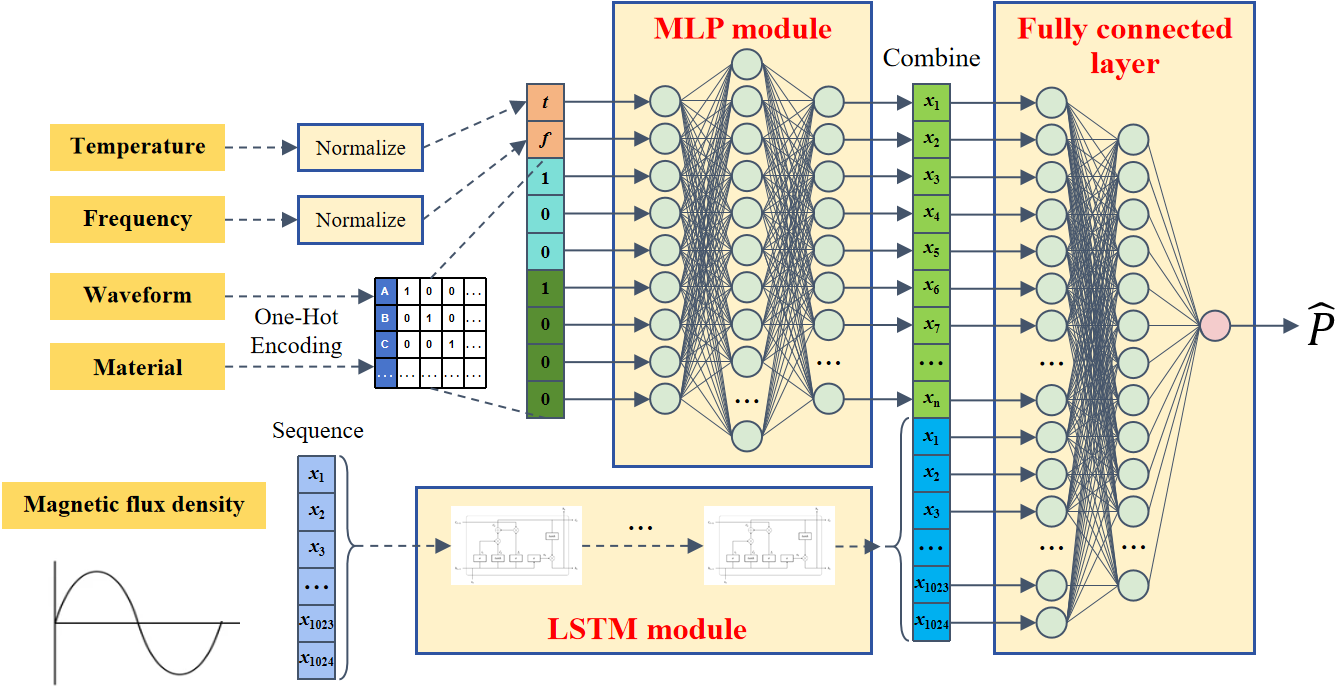}
\caption{MLP-LSTM architecture.}
\label{MLP-LSTM}
\end{figure*}

\begin{enumerate}
\item{
MLP: MLP is a basic feedforward neural network with a very simple structure, consisting of an input layer, a hidden layer, and an output layer, with neurons in each layer fully connected to neurons in the next layer. The essence of MLP feature extraction is to perform layer by layer linear transformation and nonlinear mapping on the input data, and the final output feature vector is:
\begin{equation}
{a}^{L}=\sigma ({z}^{L})=\sigma ({W}^{L}\ast {a}^{L-1}+{b}^{L})
\end{equation}
among them, $L$ is the number of hidden layers, ${a}^{L-1}$ is the output of the previous layer, ${W}^{L}$ is the weight matrix of the current layer, ${b}^{L}$ is the bias of the current layer, and $\sigma$ represents the activation function.
}

\item{
LSTM: LSTM is a type of Recurrent Neural Network (RNN) that introduces Cell State to store information and controls the information flow through gate mechanisms (input gate, forget gate, and output gate) to solve the gradient vanishing problem in RNN. The principle of LSTM is shown in the Figure \ref{LSTM}. The computation sequence of LSTM begins by inputting the current input \(x_t\) and the previous hidden state \(h_{t-1}\). Next, the forget gate \(f_t\) is computed, which decides what information to discard from the previous cell state \(C_{t-1}\). Then, the input gate \(i_t\) and the candidate cell state \(\tilde{C}_t\) are calculated, determining the new information to be added to the cell state. Following this, the current cell state \(C_t\) is updated using the outputs of the forget gate and input gate. After updating the cell state, the output gate \(o_t\) is computed to decide the current output based on the updated cell state. Finally, the current hidden state \(h_t\) is generated, which is passed to the next time step. The expression is as follows:

    \textbf{Forget Gate}:
    \begin{equation}
    f_t = \sigma(W_f \cdot [h_{t-1}, x_t] + b_f)
    \end{equation}
    \textbf{Input Gate}:
    \begin{equation}
    i_t = \sigma(W_i \cdot [h_{t-1}, x_t] + b_i)
    \end{equation}
    \textbf{Candidate Cell State}:
    \begin{equation}
    \tilde{C}_t = \tanh(W_C \cdot [h_{t-1}, x_t] + b_C)
    \end{equation}
    \textbf{Update Cell State}:
    \begin{equation}
    C_t = f_t \cdot C_{t-1} + i_t \cdot \tilde{C}_t
    \end{equation}
    \textbf{Output Gate}:
    \begin{equation}
    o_t = \sigma(W_o \cdot [h_{t-1}, x_t] + b_o)
    \end{equation}
    \textbf{Hidden State}:
    \begin{equation}
    h_t = o_t \cdot \tanh(C_t)
    \end{equation}

among them, $W$ are different weight matrices respectively, $b$ represents different bias vectors, and $\sigma$ is the activation function of Sigmod. The expression is
\begin{equation}
\sigma(x) = \frac{1}{1 + e^{-x}}
\end{equation}
}

\begin{figure}
\centering
\includegraphics[width=3.2in]{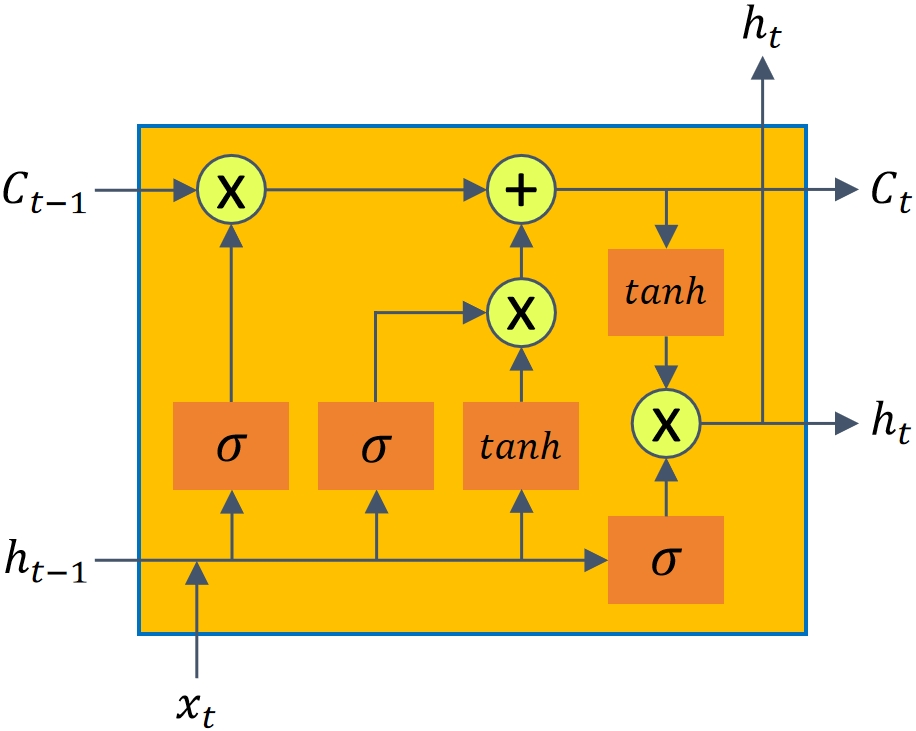}
\caption{LSTM architecture.}
\label{LSTM}
\end{figure}

\end{enumerate}

However, due to its inherent structure, LSTM is difficult to extract features from the entire sequence, so a more effective processing method is needed for complex excitation waveforms\cite{10567902}.

\subsection{Mix Neural Networks architecture}
In order to better extract features from sequences, we have proposed a Mix Neural Networks (MNN) architecture for predicting magnetic core loss. MNN combines different types of neural network models, consisting of Convolutional Neural Networks (CNN) and Fully Connected Neural Networks (FCNN), with the aim of processing datasets with complex features.

\begin{figure*}
\centering
\includegraphics[width=\textwidth]{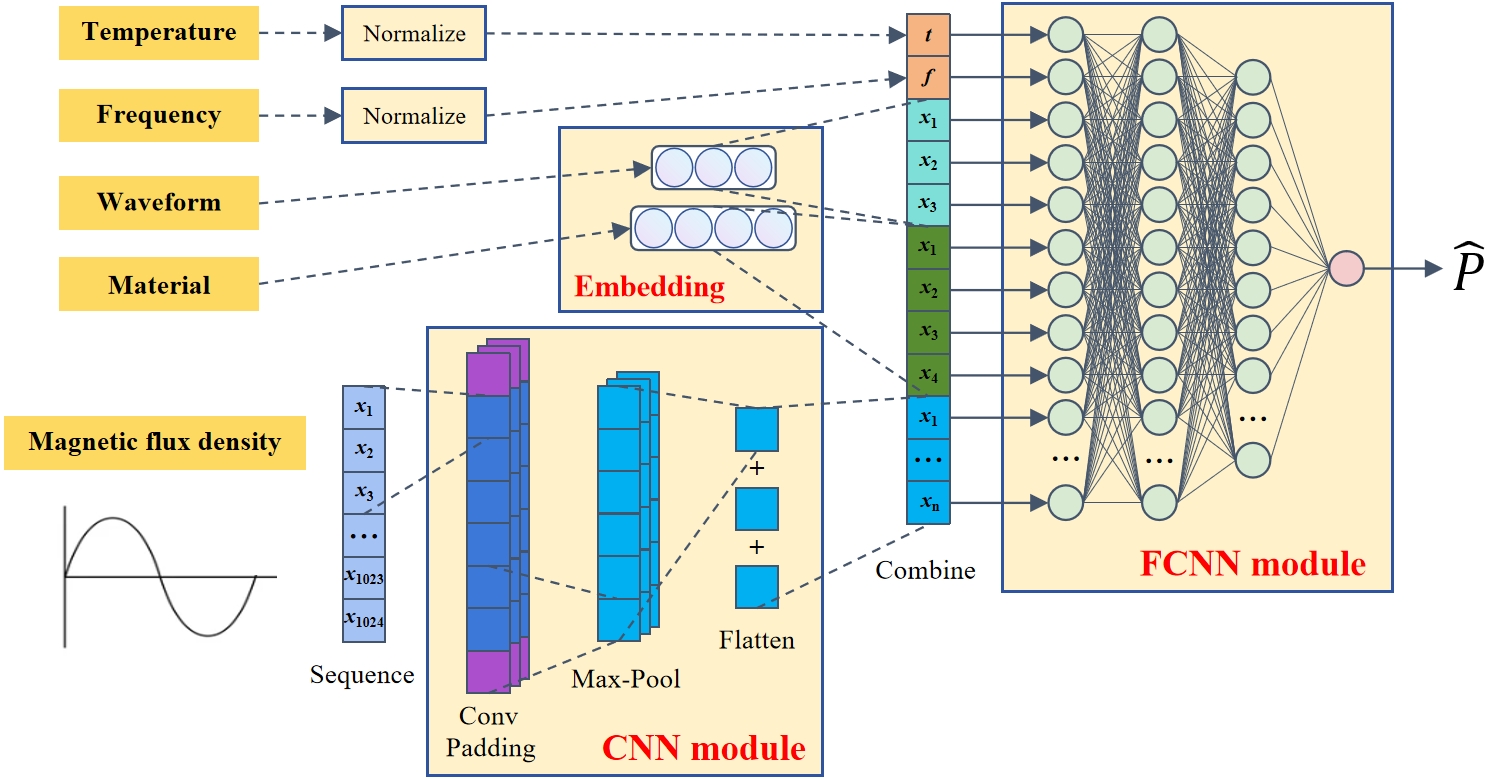}
\caption{Mix Neural Networks architecture.}
\label{cnn-rcnn}
\end{figure*}

The MNN based on CNN-FCNN can be divided into three modules: embedding layer, convolutional neural network, and fully connected neural network. Among them, the embedding layer is generally used to extract discrete features, the convolutional layer is generally used to extract sequence features, and finally concatenated in the fully connected layer and input into the neural network training. The structure of the MNN model is shown in the Figure \ref{cnn-rcnn}, and the specific principle of the model is as follows:
\begin{enumerate}
\item{
Embedding layer: converts discrete features into low dimensional continuous vectors, commonly used for processing categorical data uch as material types and waveform types.The formula can be expressed as
\begin{equation}
Embedding(x)={W}_{e}\left [ x\right ]
\end{equation}
where $W$ is the embedded weight matrix.
}
\item{
Convolutional layer: performs local perception and feature extraction on input sequences, suitable for capturing local features in waveform signals. Through multi-layer convolution, the model can extract more features. The formula for convolution is
\begin{equation}
y=Conv(x)=\sigma (W\ast x+b)
\end{equation}
among them, $W$ is the convolution kernel, $\ast$ represents the convolution operation, $b$ is the bias, and $\sigma$ represents the activation function.

}
\item{
Pooling Layer: The pooling layer performs downsampling on the input, reducing the size of the data while retaining important features. The pooling method used in this model is Max-Pooling, which selects the maximum value in the local window. The expression is
\begin{equation}
{Z}_{i}^{pool}=\max ({z}_{i},{z}_{i+1},...,{z}_{i+n-1})
\end{equation}
where $Z$ represents the maximum value of the $i$-th pooling region, and $n$ represents the window size.
}
\item{
Feature Concatenation: First, the output feature maps after convolution and pooling are flattened into a one-dimensional vector. Then, this vector is concatenated with the embedding layer vector and linear features to form a new one-dimensional vector.
}
\item
{
Fully Connected Layer: After feature concatenation, the resulting feature vector undergoes regression to predict the core loss. The predicted result is
\begin{equation}
y=\sigma ({W}_{f}x+{b}_{f})
\end{equation}
where $W$ and $b$ represent the weights and biases of the fully connected layer, respectively.
}
\end{enumerate}

\subsection{Performance indexes}
In the performance evaluation of the models, we use Mean Absolute Percentage Error (MAPE) and coefficient of determination $R^2$ as evaluation indexes. 

MAPE is a measure of the error between predicted and true values, representing the average percentage of relative error between predicted and true values. Its defined as
\begin{equation}
\text{MAPE} = \frac{1}{n} \sum_{i=1}^{n} \left| \frac{y_i - \hat{y}_i}{y_i} \right| \times 100\%
\end{equation}
where $n$ is the number of samples, $y_i$ is the true value, and $\hat{y}_i$ is the predicted value. Max APE represents the maximum error value.

$R^2$ is an indicator used to evaluate the goodness of fit of regression models, representing the explanatory power of the model on the variability of observed data. Its defined as
\begin{equation}
R^2 = 1 - \frac{\sum_{i=1}^{n} (y_i - \hat{y}_i)^2}{\sum_{i=1}^{n} (y_i - \bar{y})^2}
\end{equation}
Where $\bar{y}$ is the mean of the true values $y_i$. Normally, the value of $R^2$ is between [0,1], and the closer it is to 1, the better the model fits.

\section{Model training results}
\label{sec:3}

\subsection{Traditional model fitting}
Use the SE to perform nonlinear regression fitting on the dataset. Based on the limitations of the SE, we only selected data with sine excitation waveforms from the dataset for training, and the predicted results are shown in Table \ref{tab2}. According to the performance indexes, we can find that $R^2$ is around 0.95, MAPE is around 40, and Max APE is all greater than 130\%, indicating low prediction accuracy. 

\begin{figure}
\centering
\includegraphics[width=3.2in]{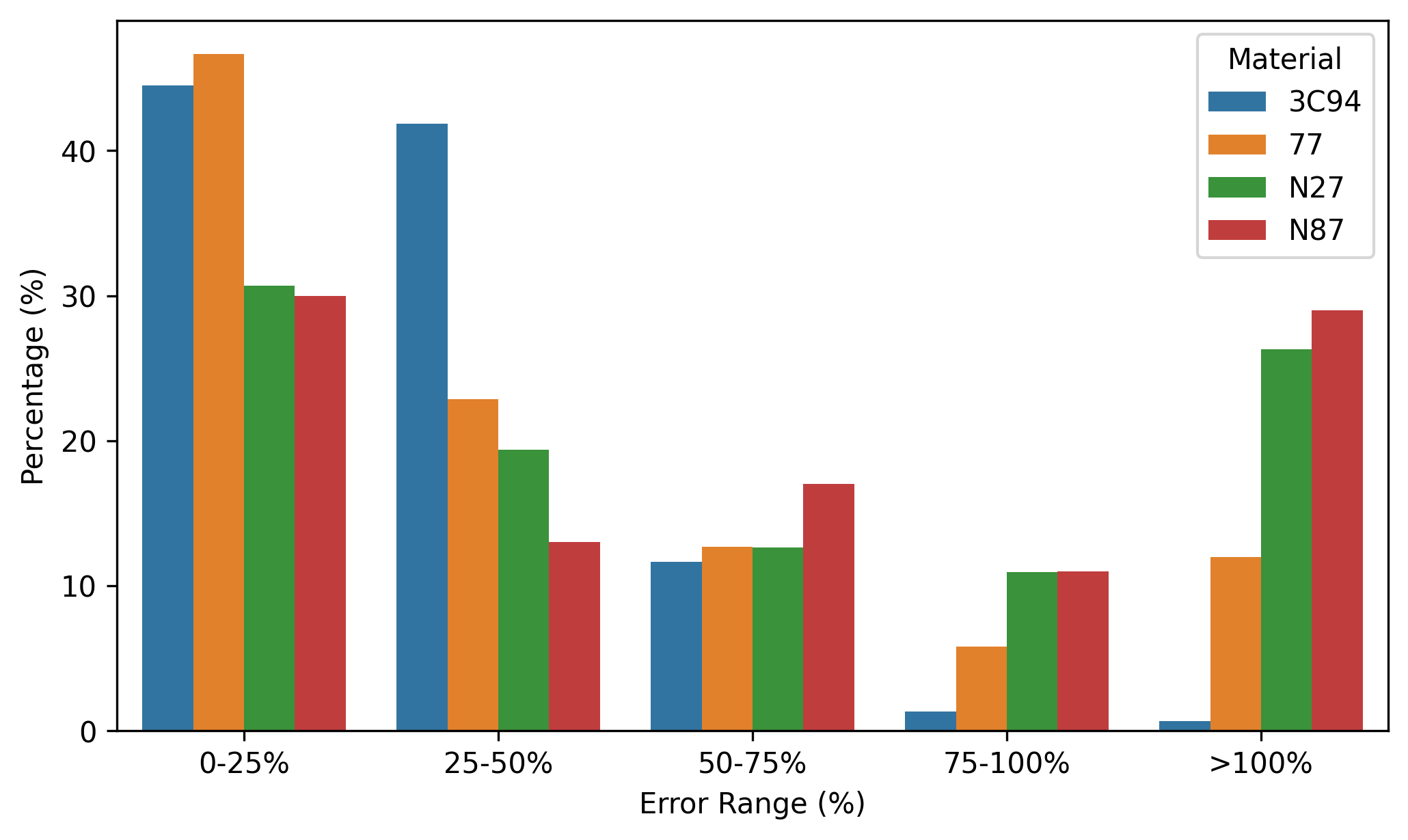}
\caption{Error distribution of iGSE prediction results.}
\label{iGSE}
\end{figure}

In order to consider the magnetic core loss under all waveforms, we continued to use the iGSE to fit the training set. The prediction results of the test-set are shown in Table \ref{tab3} and Figure \ref{iGSE}. Compared to the SE, $R^2$ have significantly decreased, and for N27 and N87, the average error is relatively large, even reaching a maximum error of 373\%. The main purpose of this correction equation is to predict all magnetic core loss without considering waveforms, but the results show a significant decrease in prediction accuracy, so the correction equation cannot fit all samples well. This is actually due to the poor fitting effect caused by a small number of features, such as both equations not considering factors such as temperature. So in the several models we will discuss next, we have incorporated the influence of temperature on magnetic core loss. 

\begin{table}
\centering
\caption{The coefficient and performance of SE}
\label{tab2}
\renewcommand{\arraystretch}{1.2} 
\begin{tabular}{ccccc}
\hline
Material  & Coefficients($k$, $a$, $b$)  & MAPE(\%) & Max APE(\%) & $R^2$   \\ \hline
3C94      & {[}1.500, 1.430, 2.471{]} & 35.662  & 157.804  & 0.945 \\
77        & {[}0.561, 1.513, 2.321{]} & 47.063  & 260.443  & 0.940 \\
N27       & {[}0.971, 1.490, 2.390{]} & 39.740  & 182.703  & 0.953 \\
N87       & {[}0.400, 1.578, 2.453{]} & 33.102  & 133.967  & 0.956 \\ \hline
\end{tabular}
\end{table}

\begin{table}
\centering
\caption{The coefficient and performance of iGSE}
\label{tab3}
\renewcommand{\arraystretch}{1.2} 
\begin{tabular}{ccccc}
\hline
Material  & Coefficients($k$, $a$, $b$)  & MAPE(\%) & Max APE(\%) & $R^2$   \\ \hline
3C94      & {[}0.024, 1.590, 2.204{]} & 29.268  & 146.044  & 0.872 \\
77        & {[}0.042, 1.570, 2.224{]} & 44.279  & 263.109  & 0.951 \\
N27       & {[}0.138, 1.430, 2.265{]} & 71.753  & 373.009  & 0.903 \\
N87       & {[}0.306, 1.450, 2.266{]} & 72.861  & 222.059  & 0.946 \\ \hline
\end{tabular}
\end{table}

\subsection{Machine learning model training}
By analyzing the original dataset, it can be found that it is composed of linear data (temperature, frequency), nonlinear data (material type, waveform type), and sequences (magnetic flux density). There are two ways to process such complex dataset: by performing feature dimensionality reduction on the sequence, the reduced features are regressed with nonlinear and linear data, or by packaging the sequence, linear data, and nonlinear data into a computer to automatically extract features.

Based on machine learning, we use Random Forest and XGBoost for training. The advantage over traditional models is that there is no need to classify and discuss magnetic core categories again; Secondly, due to more features, the relative accuracy will also be higher. Of course, the disadvantage is that the more features there are, the larger the data volume and the more complex the model will be. Two models are suitable for processing nonlinear data, but require manual feature extraction. In feature extraction, we extracted 33 features based on sequences. A total of 37 features were introduced in the model training, including sequence features and other linear and nonlinear features. The hyperparameter settings for Random Forest and XGBoost are shown in Table \ref{tab4}, respectively. Mean-square error (MSE) is used as the loss function, and its calculation formula is
\begin{equation}
\text{MSE} = \frac{1}{n} \sum_{i=1}^{n} (y_i - \hat{y}_i)^2
\end{equation}
where $n$ is the number of samples, $y_i$ is the true value, and $\hat{y}_i$ is the predicted value.

\begin{figure}[h]
\centering
\includegraphics[width=3.2in]{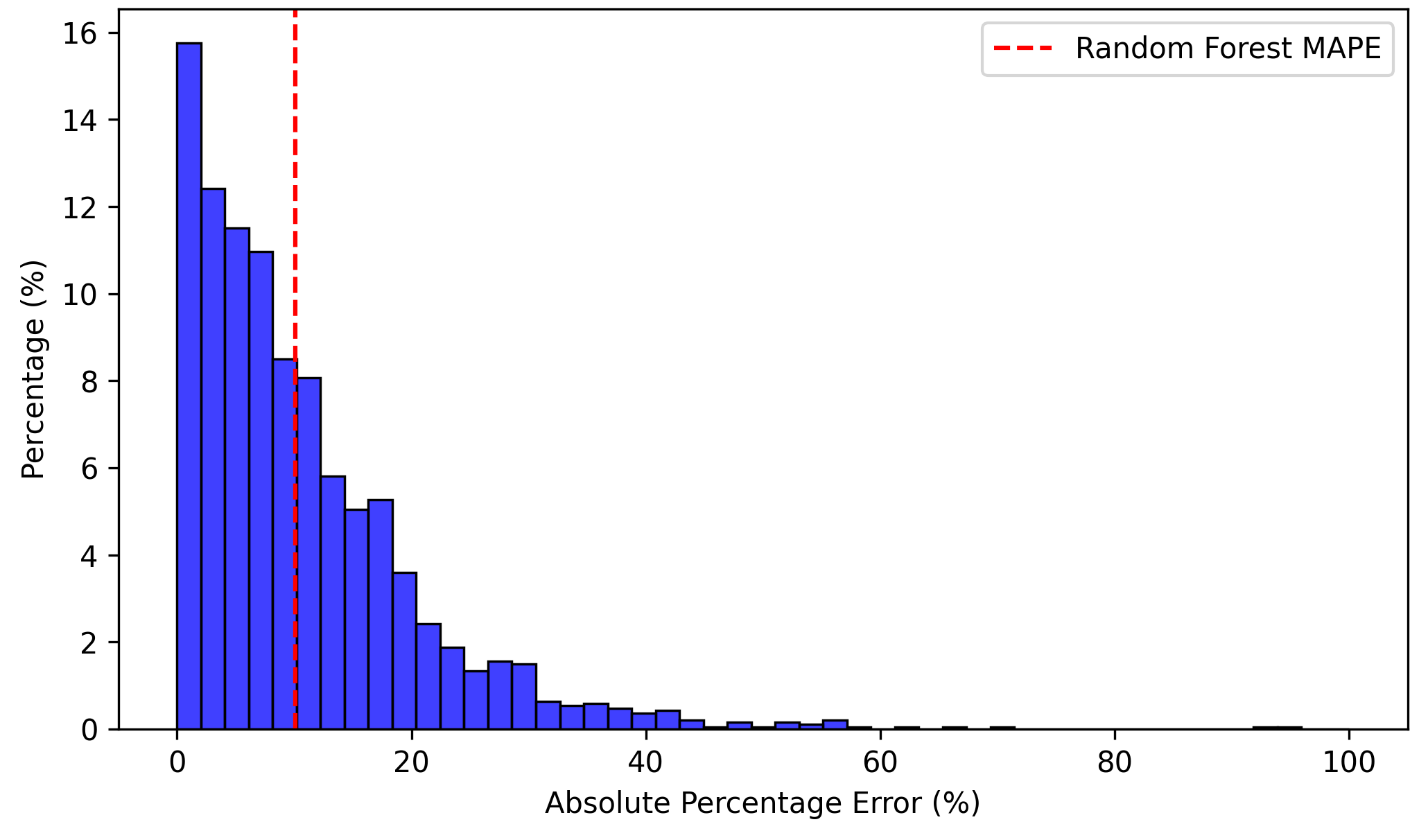}
\caption{Error histogram of Random Forest prediction results.}
\label{fig2}
\end{figure}

\begin{figure}
\centering
\includegraphics[width=3.2in]{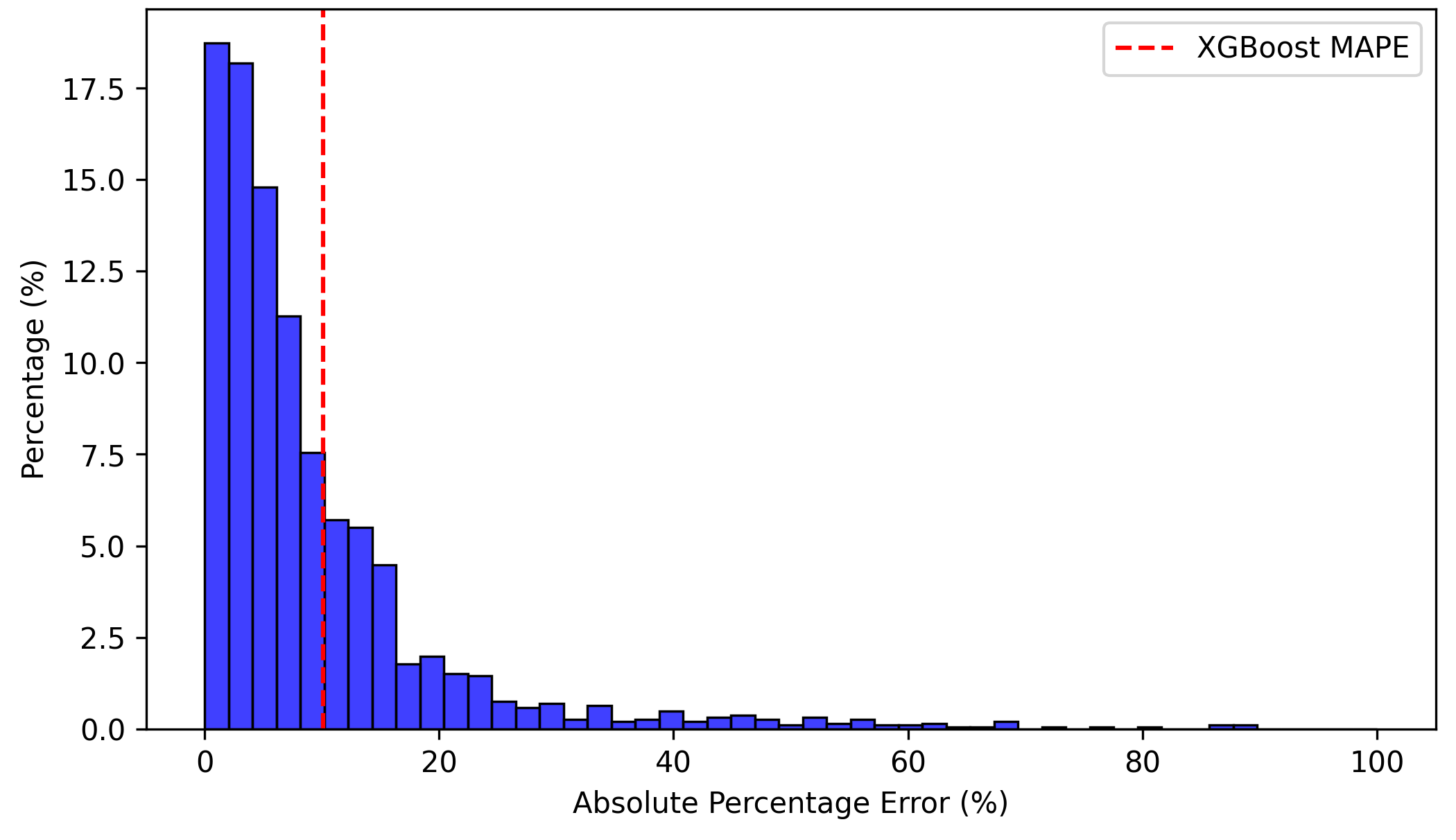}
\caption{Error histogram of XGBoost prediction results.}
\label{fig3}
\end{figure}

The prediction error distributions of the two models are shown in Figure \ref{fig2} and \ref{fig3}. Machine learning models can already make up for the limitations of traditional models very well, and from the results, the prediction accuracy has been greatly improved. Machine learning models can already effectively compensate for the limitations of traditional models, and the results show a significant improvement in prediction accuracy. It can be observed that the majority of predicted values have an error distribution within 10\%, with only about 15\% of the total predicted values having an error of over 20\%.

\begin{table}[h]
\centering
\caption{machine learning models hyperparameter selection}
\label{tab4}
\renewcommand{\arraystretch}{1.2} 
\begin{tabular}{ccc}
\hline
Model                          & Hyperparameter          & Value            \\ \hline
\multirow{4}{*}{Random Forest} & n\_estimators           & 100              \\
                               & criterion               & MSE              \\
                               & min\_samples\_split     & 2                \\
                               & min\_samples\_leaf      & 1                \\ \hline
\multirow{5}{*}{XGBoost}       & num\_boost\_round       & 10000            \\
                               & early\_stopping\_rounds & 50               \\
                               & max\_depth              & 6                \\
                               & learning\_rate          & 0.01             \\
                               & objective               & reg:squarederror \\ \hline
\end{tabular}
\end{table}

\subsection{Deep learning model training}
\begin{figure}
\centering
\includegraphics[width=3.2in]{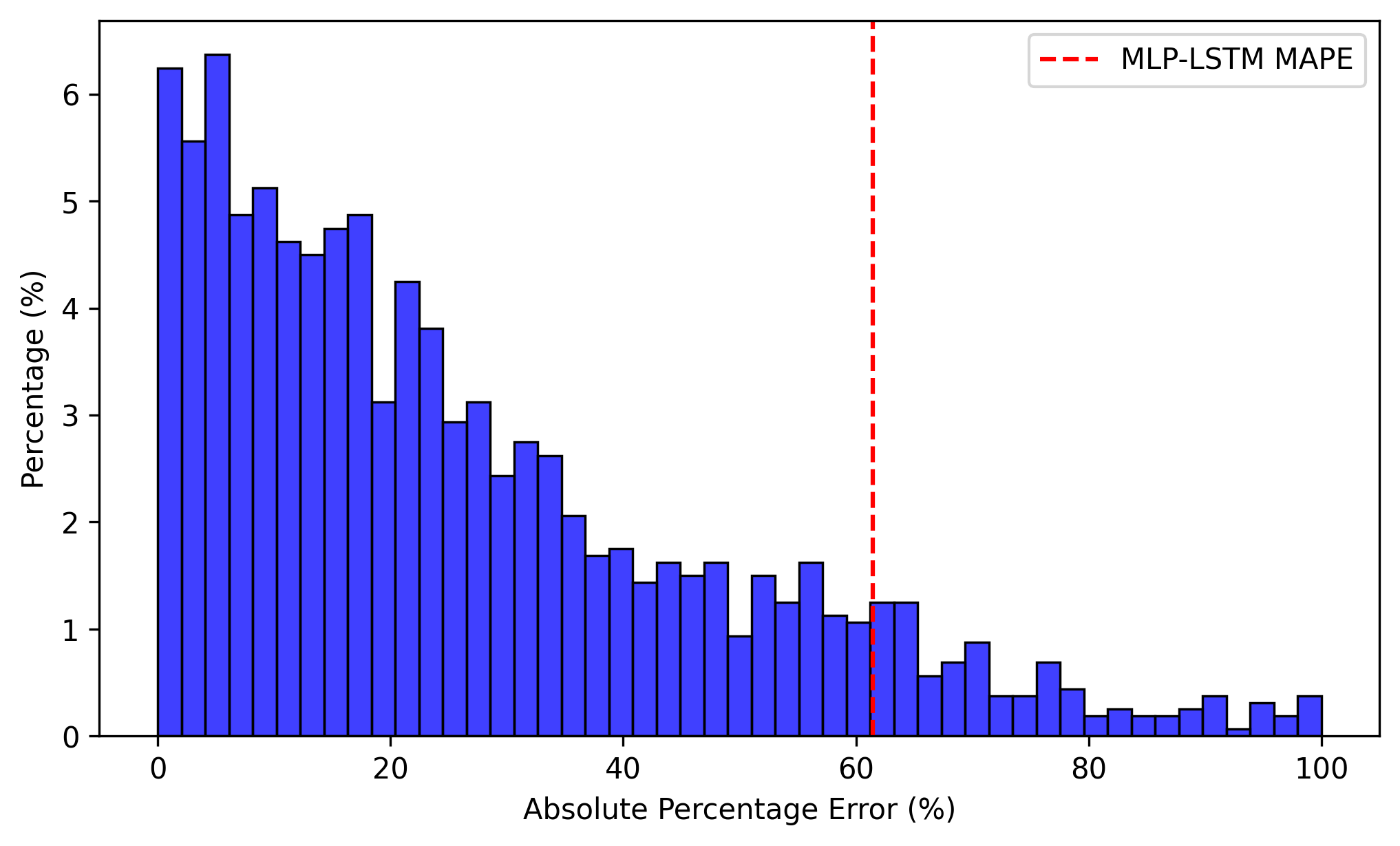}
\caption{Error histogram of MLP-LSTM Forest prediction results.}
\label{fig4}
\end{figure}

\begin{figure}
\centering
\includegraphics[width=3.2in]{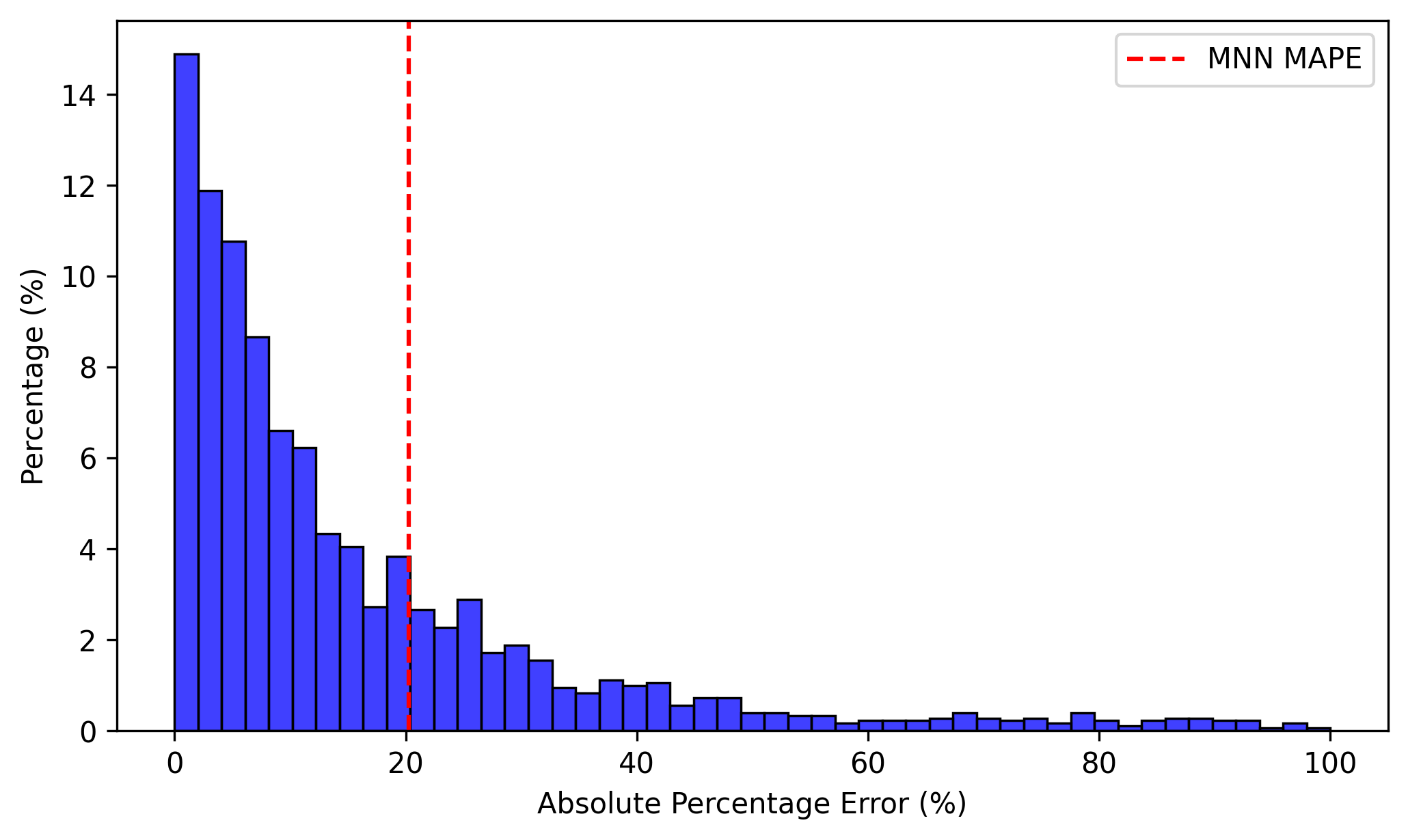}
\caption{Error histogram of MNN prediction results.}
\label{fig5}
\end{figure}

Traditional machine learning models usually do not have the ability to simultaneously regress three types of data. Based on the complex nonlinear relationships between features, we use the MLP-LSTM model and our newly proposed MNN model. For deep learning models, features will be automatically extracted from sequences through operations such as convolution, so we do not need to perform any other operations on the dataset. The hyperparameter settings for two types of deep learning are shown in Table \ref{tab5}, where the calculation formula for the activation function Rectified Linear Unit (ReLu) is  
\begin{equation}
f(x)=\max (0, x)
\end{equation}
and the prediction error distributions for the test set of the two models are shown in Figure \ref{fig4} and \ref{fig5}. It can be clearly observed that the prediction results of MLP-LSTM are significantly worse than MNN, and worse than Random Forest and XGBoost.

\begin{table}
\centering
\caption{MLP-LSTM hyperparameter selection}
\label{tab5}
\renewcommand{\arraystretch}{1.2} 
\begin{tabular}{ccc}
\hline
Model                      & Hyperparameter                        & Value \\ \hline
\multirow{5}{*}{MLP}       & number of layers                      & 2     \\
                           & number of neurons in the first layer  & 128   \\
                           & number of neurons in the second layer & 64    \\
                           & activation function                   & ReLU  \\
                           & dropout rate                          & 0.5   \\ \hline
\multirow{3}{*}{LSTM}      & input dimension                       & 1024  \\
                           & Number of Hidden Units                & 128   \\
                           & Number of Layers                      & 2     \\ \hline
\multirow{4}{*}{Final-FCL} & Number of Layers                      & 1     \\
                           & number of neurons                     & 128   \\
                           & activation function                   & ReLU  \\
                           & dropout rate                          & 0.5   \\ \hline
\multirow{5}{*}{MLP-LSTM}  & Epochs                                & 10000 \\
                           & Learning Rate                         & 0.001 \\
                           & optimizer                             & Adam  \\
                           & loss function                         & MSE   \\
                           & patience                              & 50    \\ \hline
\end{tabular}
\end{table}

\begin{table}
\centering
\caption{MNN hyperparameter selection}
\label{tab6}
\renewcommand{\arraystretch}{1.2} 
\begin{tabular}{cccc}
\hline
Model                & \multicolumn{2}{c}{Hyperparameter}                            & Value \\ \hline
\multirow{9}{*}{CNN}  & \multirow{4}{*}{first convolutional layer}   & in\_channels & 1    \\
                     &                                               & out\_channels & 16    \\
                     &                                               & kernel\_size  & 3     \\
                     &                                               & padding       & 1     \\ \cline{2-4} 
                     & \multirow{4}{*}{second convolutional layer}   & in\_channels  & 16    \\
                     &                                               & out\_channels & 32    \\
                     &                                               & kernel\_size  & 3     \\
                     &                                               & padding       & 1     \\ \cline{2-4} 
                     & pooling layer                                 & kernel\_size  & 2     \\ \hline
\multirow{7}{*}{FCNN} & \multirow{2}{*}{first fully connected layer} & in\_features & 8201 \\
                     &                                               & out\_features & 128   \\ \cline{2-4} 
                     & \multirow{2}{*}{second fully connected layer} & in\_features  & 128   \\
                     &                                               & out\_features & 64    \\ \cline{2-4} 
                     & \multirow{2}{*}{output fully connected layer} & in\_features  & 64    \\
                     &                                               & out\_features & 1     \\ \cline{2-4} 
                     & \multicolumn{2}{c}{activation function}                       & ReLU  \\ \hline
\multirow{5}{*}{MNN} & \multicolumn{2}{c}{Epochs}                                    & 1000  \\
                     & \multicolumn{2}{c}{Learning Rate}                             & 0.001 \\
                     & \multicolumn{2}{c}{optimizer}                                 & Adam  \\
                     & \multicolumn{2}{c}{loss function}                             & MSE   \\
                     & \multicolumn{2}{c}{patience}                                  & 20    \\ \hline
\end{tabular}
\end{table}

\subsection{Analysis and Discussion}

Analyzing the four models, the performance indexes are shown in Table \ref{tab6}: For MSE, XGBoost and MNN are not on the same order of magnitude as the other two models, indicating that their error is much smaller than the other two models; For both MAPE and Max APE, MLP-LSTM is the model with the worst prediction performance, and deep learning is worse than machine learning models, with a gap of at least twice as much; The $R^2$ of all four models is high, but MLP-LSTM is the lowest. Overall, MLP-LSTM has the worst performance. Although Random Forest has the smallest Max APE, its MSE is too large. Therefore, XGBoost and MNN have stronger predictive abilities than the other two models.

\begin{table}
\centering
\caption{The coefficient and performance of all models}
\label{tab6}
\renewcommand{\arraystretch}{1.2} 
\begin{tabular}{ccccc}
\hline
Model          & MSE        & MAPE(\%) & Max APE(\%) & $R^2$ \\ \hline
Random Forest  & 2261992560 & 10.89    & 108.025     & 0.981 \\
XGBoost        & 646750656  & 10.10    & 206.765     & 0.994 \\
MLP-LSTM       & 6423268352 & 61.46    & 1452.910    & 0.945 \\
MNN            & 593100352  & 20.25    & 498.786     & 0.995 \\
Hybrid model   & 334314144  & 12.09    & 288.215     & 0.997 \\
\hline
\end{tabular}
\end{table}

\begin{figure}
\centering
\includegraphics[width=3.2in]{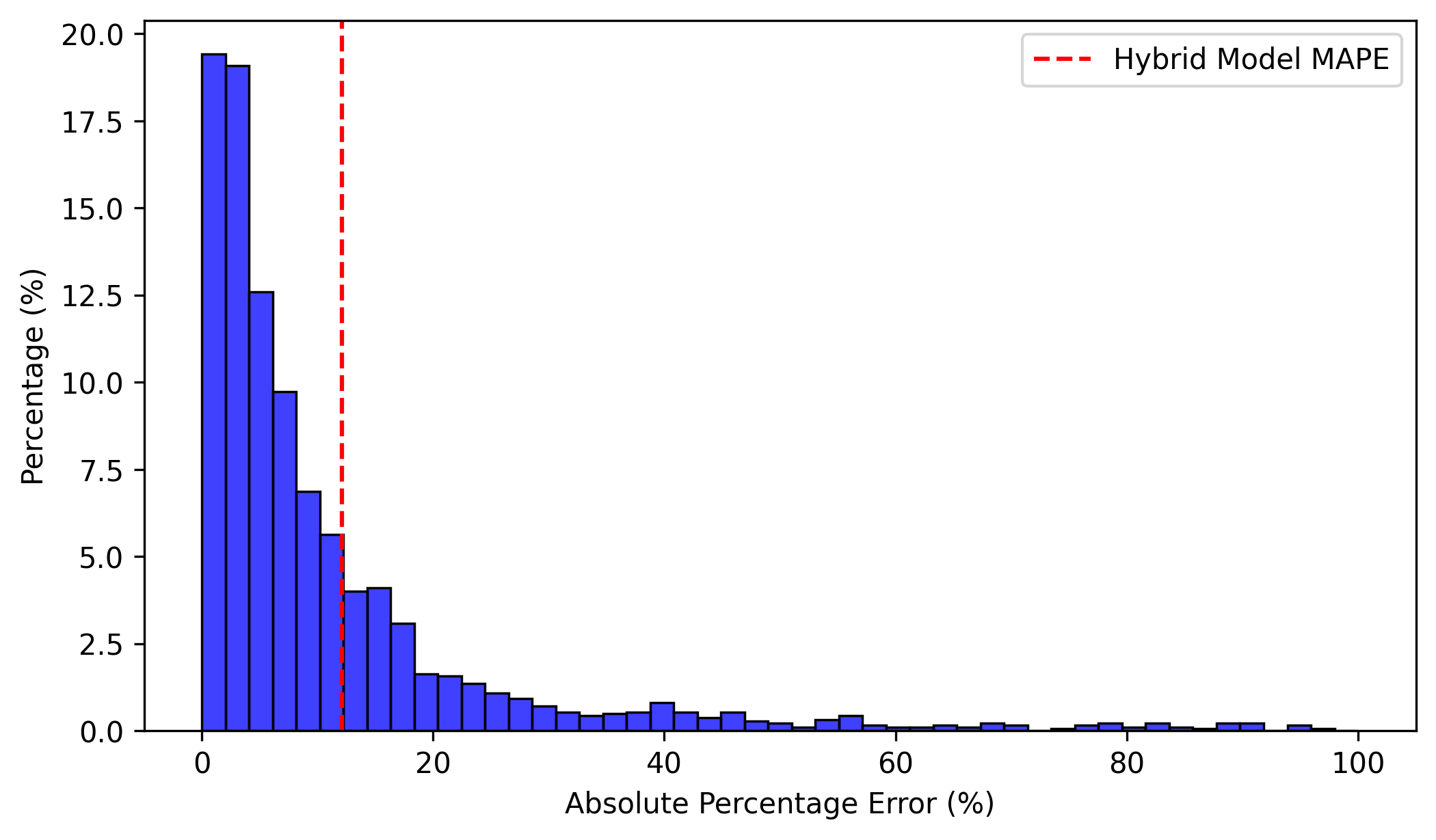}
\caption{Error histogram of hybrid model prediction results.}
\label{fig6}
\end{figure}

\begin{figure}
\centering
\includegraphics[width=3.2in]{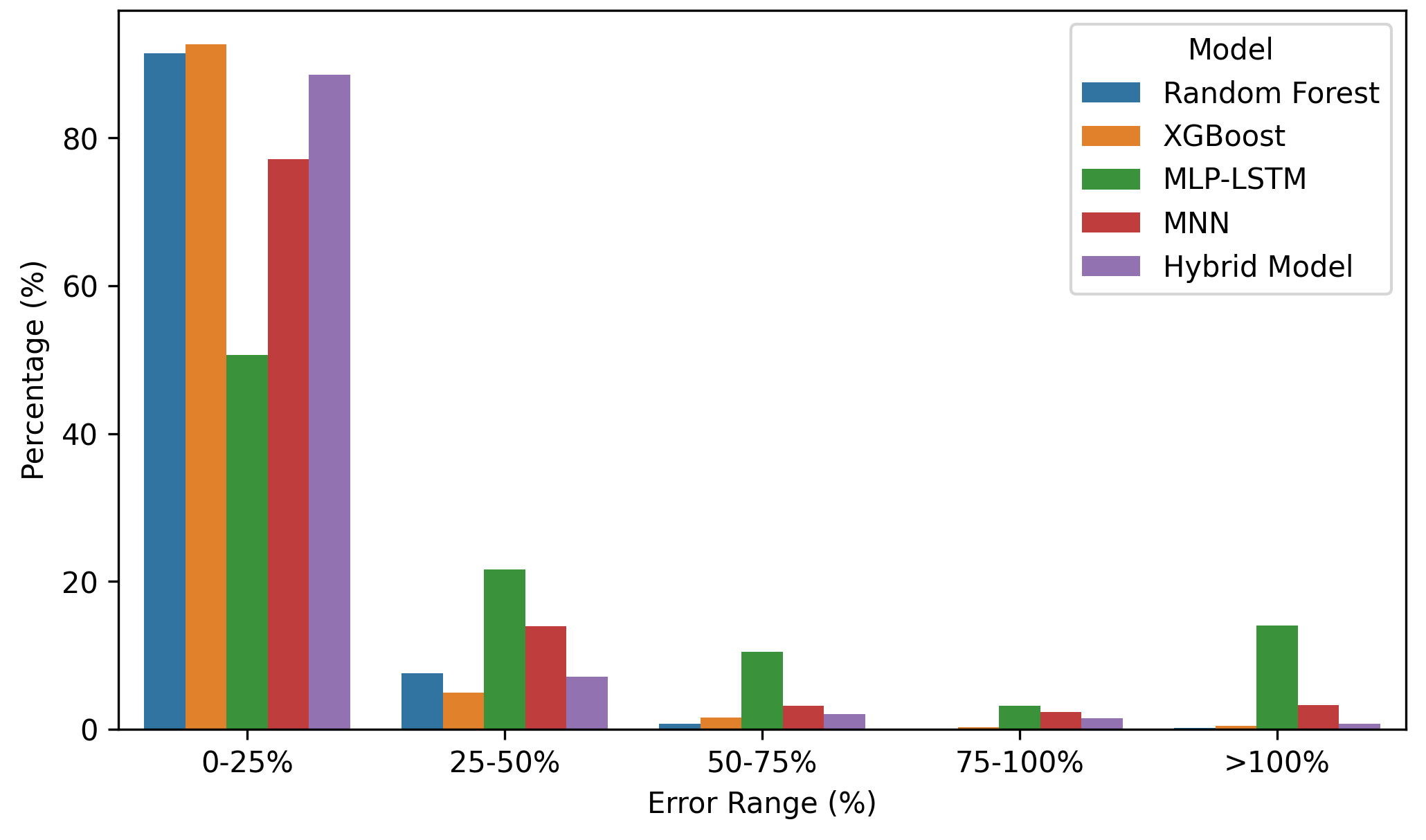}
\caption{Error distribution of all models prediction results.}
\label{fig7}
\end{figure}

\begin{figure}
\centering
\includegraphics[width=3.2in]{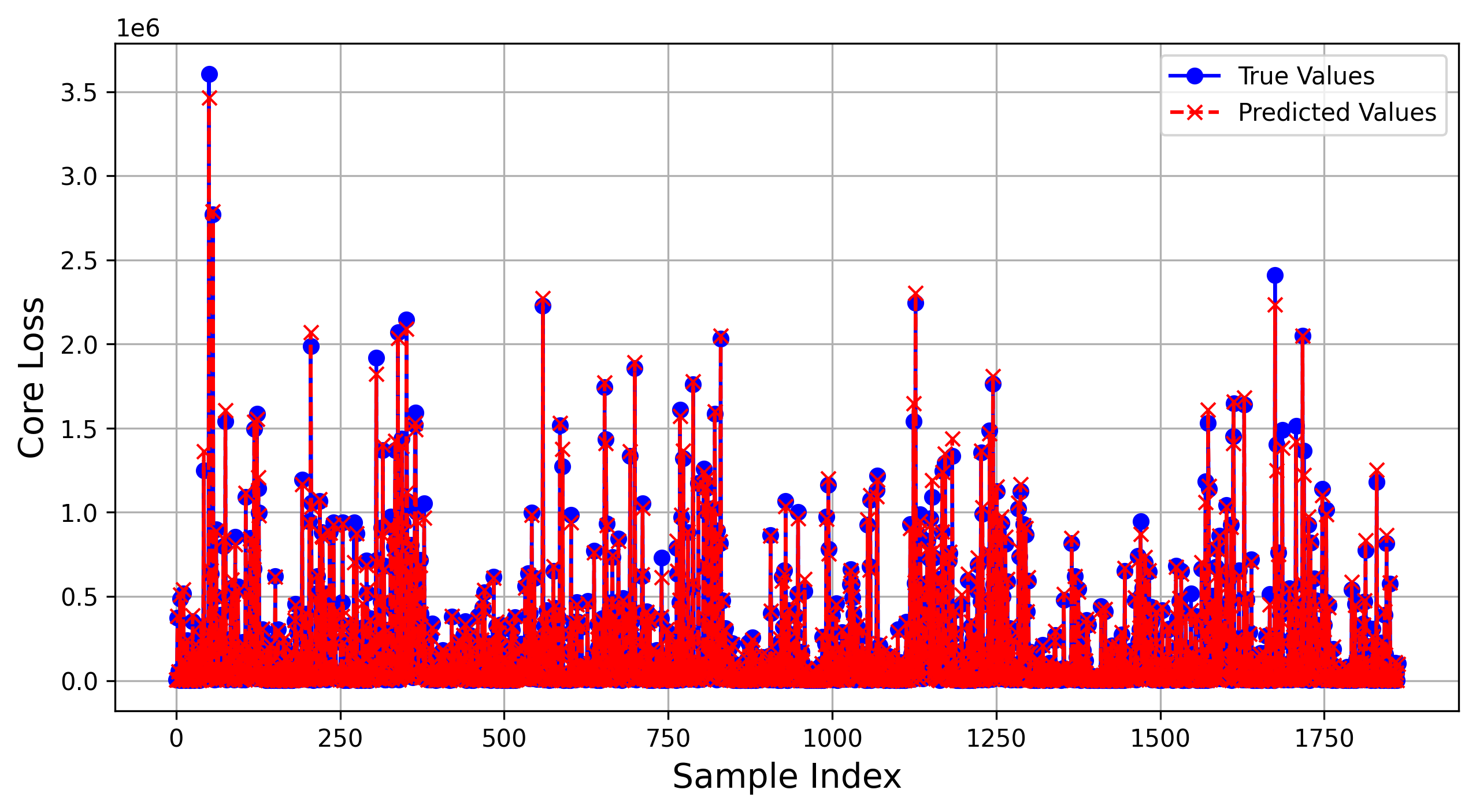}
\caption{Comparison between the predicted results of the hybrid model and the actual values.}
\label{fig8}
\end{figure}

\begin{figure}
\centering
\includegraphics[width=3.2in]{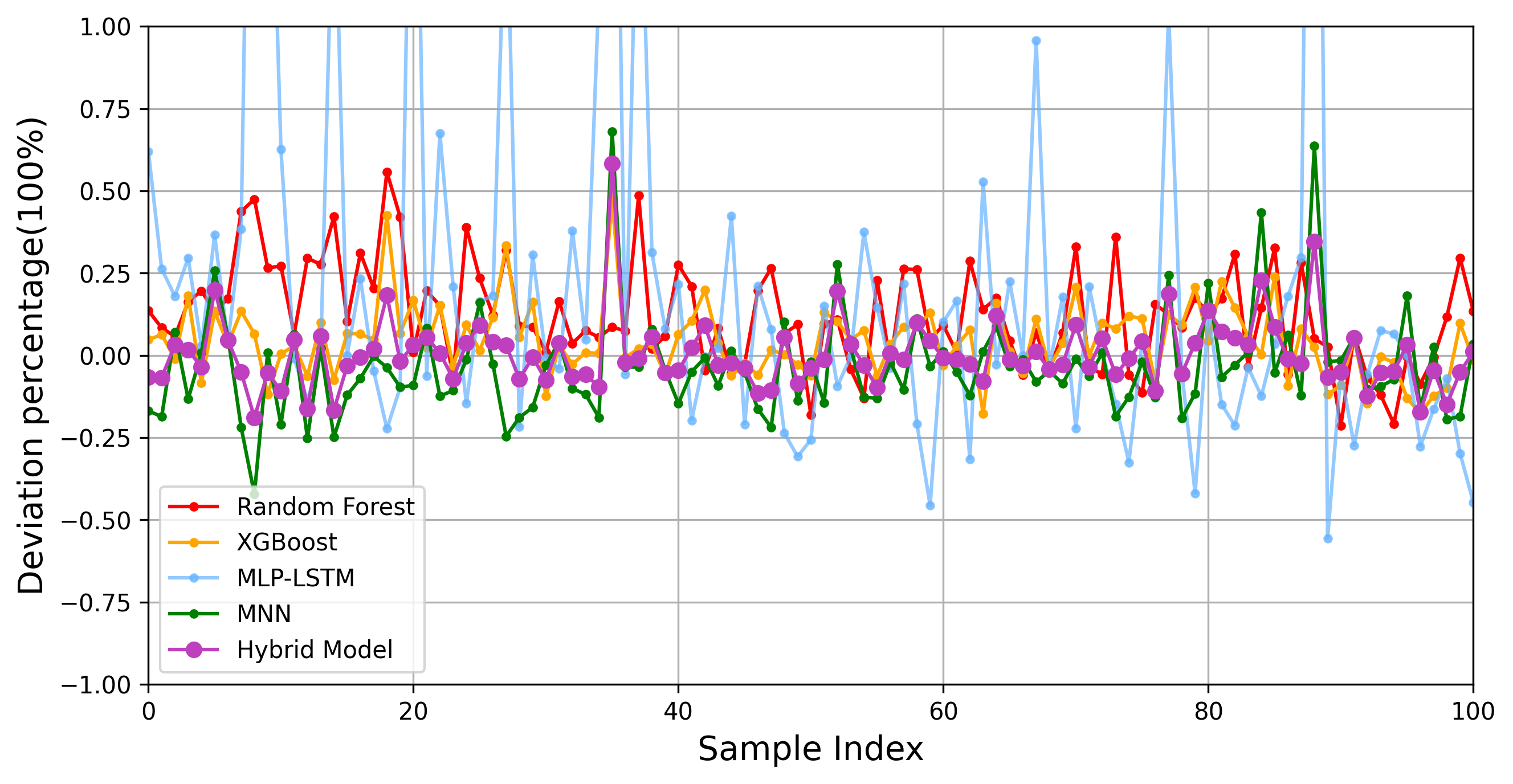}
\caption{Percentage of residual between predicted and actual values of all models.}
\label{fig9}
\end{figure}

XGBoost and MNN have their own advantages in MSE and MAPE. In order to further reduce errors, we conducted weighted analysis on these two models. Adopt
\begin{equation}
\widehat{P}_{\text{HM}}=W_1\times \widehat{P}_{\text{XGBoost}} + W_2\times \widehat{P}_{\text{MNN}}
\end{equation}
weighted predicted values can be obtained. By conducting a grid search to find the optimal weight, with the goal of finding the minimum MSE, we finally obtained $W_1=0.4764$ and $W_1=0.5236$. Under weighted conditions, we obtained the minimum MSE value as shown in Table \ref{tab6}. While MSE decreased, MAPE decreased significantly compared to MNN, with $R^2$ reaching 0.997. Figure \ref{fig6} shows the error distribution of the hybrid model on the test set, which is similar to XGBoost and better than MNN and MLP-LSTM. The error distribution statistics of all models are shown in Figure \ref{fig7}. It can be found that over 80\% of the sample errors in our proposed hybrid model are less than 25\%, indicating that this weighted model combines the advantages of machine learning and deep learning. The MSE of the hybrid model is at least 43\% better than that of deep learning and maintains the same performance as machine learning on MAPE. The final prediction result is shown in Figure \ref{fig8}, and it can be found that the predicted result is basically consistent with the actual result. Figure \ref{fig9} shows the comparison of the deviation between the predicted values and the actual values of 100 randomly selected samples, which can more clearly reflect the error fluctuation of the predicted values. It can be found that except for two prediction results with a deviation of over 25\%, the hybrid model has prediction errors within the 25\% range, and the fluctuation amplitude is smaller than the other four models.

All the machine learning and deep learning models we use have higher prediction accuracy than traditional models. By comparing some of the latest research results, it can also be found that the accuracy of our proposed model does not significantly lag behind some existing models under different materials, indicating that machine learning and deep learning models have achieved prediction of magnetic core loss for various materials. In fact, there are many factors that affect the magnetic core loss, such as temperature. However, through data-driven modeling, we only need to incorporate temperature as a feature during the training process to obtain a model with different temperature prediction capabilities. In fact, the model we proposed has already effectively solved the problem of inaccurate prediction at different temperatures. We trained the model on four temperature points (25, 50, 70, 90) in the dataset as linear data. The trained model can not only predict the core loss at these four temperature points, but also predict the entire temperature range. This demonstrates the powerful generalization ability of machine learning and deep learning models. This indicates that machine learning and deep learning models can achieve a magnetic core loss prediction model that can span different material types and operating modes. In practical applications, such a model will greatly improve the accuracy and efficiency of magnetic component design, save time and cost of repeated experiments in design, and provide assistance for the further development of power electronics technology.

\section{conclusion}
\label{sec:4}

This article proposes a model that combines machine learning and deep learning to predict magnetic core loss. We mainly discussed the predictive performance of four models on the dataset, and finally found that XGBoost and our newly proposed MNN model had significantly better generalization ability than the other two models, and we calculated the weighted prediction results and found that the accuracy was higher.

By utilizing machine learning and deep learning, we can now predict the core loss of four different materials with good accuracy using only one model. This eliminates the time and cost of training models separately for a single material or under a single condition. In practical applications, our proposed method provides a way to establish a universal model, proving that a single model can be used to predict magnetic core loss under different materials, frequencies, waveforms, and other characteristics through the combination of machine learning and deep learning.
\balance

\bibliographystyle{IEEEtran}

\begin{thebibliography}{10}
\providecommand{\url}[1]{#1}
\csname url@samestyle\endcsname
\providecommand{\newblock}{\relax}
\providecommand{\bibinfo}[2]{#2}
\providecommand{\BIBentrySTDinterwordspacing}{\spaceskip=0pt\relax}
\providecommand{\BIBentryALTinterwordstretchfactor}{4}
\providecommand{\BIBentryALTinterwordspacing}{\spaceskip=\fontdimen2\font plus
\BIBentryALTinterwordstretchfactor\fontdimen3\font minus \fontdimen4\font\relax}
\providecommand{\BIBforeignlanguage}[2]{{%
\expandafter\ifx\csname l@#1\endcsname\relax
\typeout{** WARNING: IEEEtran.bst: No hyphenation pattern has been}%
\typeout{** loaded for the language `#1'. Using the pattern for}%
\typeout{** the default language instead.}%
\else
\language=\csname l@#1\endcsname
\fi
#2}}
\providecommand{\BIBdecl}{\relax}
\BIBdecl

\bibitem{7570571}
C.~R. Sullivan, B.~A. Reese, A.~L.~F. Stein, and P.~A. Kyaw, ``On size and magnetics: Why small efficient power inductors are rare,'' in \emph{2016 International Symposium on 3D Power Electronics Integration and Manufacturing (3D-PEIM)}, 2016, pp. 1--23.

\bibitem{278656}
W.~Roshen, ``Ferrite core loss for power magnetic components design,'' \emph{IEEE Transactions on Magnetics}, vol.~27, no.~6, pp. 4407--4415, 1991.

\bibitem{9477291}
Z.~Yi, H.~Liu, and K.~Sun, ``A novel calorimetric loss measurement method for high-power high-frequency transformer based on surface temperature measurment,'' in \emph{2020 4th International Conference on HVDC (HVDC)}, 2020, pp. 645--648.

\bibitem{4802625}
D.~J. Perreault, J.~Hu, J.~M. Rivas, Y.~Han, O.~Leitermann, R.~C. Pilawa-Podgurski, A.~Sagneri, and C.~R. Sullivan, ``Opportunities and challenges in very high frequency power conversion,'' in \emph{2009 Twenty-Fourth Annual IEEE Applied Power Electronics Conference and Exposition}, 2009, pp. 1--14.

\bibitem{10700105}
C.~Li, W.~Qin, Z.~Wang, X.~Ma, W.~Wang, and M.~Cheng, ``Core loss model for non-sinusoidal excitataions based on vector magnetic circuit theory,'' in \emph{2024 International Conference on Electrical Machines (ICEM)}, 2024, pp. 1--6.

\bibitem{7395385}
S.~Li, Y.~Li, W.~Choi, and B.~Sarlioglu, ``High-speed electric machines: Challenges and design considerations,'' \emph{IEEE Transactions on Transportation Electrification}, vol.~2, no.~1, pp. 2--13, 2016.

\bibitem{10707354}
C.~Li, M.~Cheng, W.~Qin, Z.~Wang, X.~Ma, and W.~Wang, ``Analytical loss model for magnetic cores based on vector magnetic circuit theory,'' \emph{IEEE Open Journal of Power Electronics}, pp. 1--11, 2024.

\bibitem{5570437}
C.~P. Steinmetz, ``On the law of hysteresis,'' \emph{Transactions of the American Institute of Electrical Engineers}, vol.~IX, no.~1, pp. 1--64, 1892.

\bibitem{1196712}
K.~Venkatachalam, C.~Sullivan, T.~Abdallah, and H.~Tacca, ``Accurate prediction of ferrite core loss with nonsinusoidal waveforms using only steinmetz parameters,'' in \emph{2002 IEEE Workshop on Computers in Power Electronics, 2002. Proceedings.}, 2002, pp. 36--41.

\bibitem{955931}
J.~Li, T.~Abdallah, and C.~Sullivan, ``Improved calculation of core loss with nonsinusoidal waveforms,'' in \emph{Conference Record of the 2001 IEEE Industry Applications Conference. 36th IAS Annual Meeting (Cat. No.01CH37248)}, vol.~4, 2001, pp. 2203--2210 vol.4.

\bibitem{5955126}
J.~Muhlethaler, J.~Biela, J.~W. Kolar, and A.~Ecklebe, ``Improved core-loss calculation for magnetic components employed in power electronic systems,'' \emph{IEEE Transactions on Power Electronics}, vol.~27, no.~2, pp. 964--973, 2012.

\bibitem{936396}
J.~Reinert, A.~Brockmeyer, and R.~De~Doncker, ``Calculation of losses in ferro- and ferrimagnetic materials based on the modified steinmetz equation,'' \emph{IEEE Transactions on Industry Applications}, vol.~37, no.~4, pp. 1055--1061, 2001.

\bibitem{548774}
M.~Albach, T.~Durbaum, and A.~Brockmeyer, ``Calculating core losses in transformers for arbitrary magnetizing currents a comparison of different approaches,'' in \emph{PESC Record. 27th Annual IEEE Power Electronics Specialists Conference}, vol.~2, 1996, pp. 1463--1468 vol.2.

\bibitem{5367762}
K.~Passadis and G.~Loizos, ``Core power losses estimation of wound core distribution transformers with support vector machines,'' in \emph{2009 16th International Conference on Systems, Signals and Image Processing}, 2009, pp. 1--4.

\bibitem{10608787}
L.~Solimene, C.~S. Ragusa, A.~Giuffrida, N.~Lombardo, F.~Marmello, S.~Morra, and M.~Pasquale, ``A hybrid data-driven approach in magnetic core loss modeling for power electronics applications,'' in \emph{2024 IEEE 22nd Mediterranean Electrotechnical Conference (MELECON)}, 2024, pp. 1356--1361.

\bibitem{10615222}
M.~Choi, S.~Park, E.~Jang, M.~Ouk, K.~Park, S.~Lee, and G.~Noh, ``Fabrication-specific simulation of mn-zn ferrite core-loss for machine learning-based surrogate modeling with limited experimental data,'' \emph{IEEE Transactions on Power Electronics}, pp. 1--12, 2024.

\bibitem{10315145}
J.~Deng, W.~Wang, Z.~Ning, P.~Venugopal, J.~Popovic, and G.~Rietveld, ``High-frequency core loss modeling based on knowledge-aware artificial neural network,'' \emph{IEEE Transactions on Power Electronics}, vol.~39, no.~2, pp. 1968--1973, 2024.

\bibitem{10567902}
B.~Su, K.~Sun, M.~Yang, Y.~Xiao, K.~Zhang, and B.~Liu, ``A core loss estimation method based on data-driven technology with multi-head attention mechanism,'' in \emph{2024 IEEE 10th International Power Electronics and Motion Control Conference (IPEMC2024-ECCE Asia)}, 2024, pp. 3992--3996.

\bibitem{9907097}
X.~Shen, H.~Wouters, and W.~Martinez, ``Deep neural network for magnetic core loss estimation using the magnet experimental database,'' in \emph{2022 24th European Conference on Power Electronics and Applications (EPE'22 ECCE Europe)}, 2022, pp. 1--8.

\bibitem{sapkota2024deep}
D.~B. Sapkota, P.~Neupane, M.~Joshi \emph{et~al.}, ``Deep learning model for enhanced power loss prediction in the frequency domain for magnetic materials,'' \emph{IET Power Electronics}, vol.~17, no.~12, pp. 1--8, 2024.

\bibitem{10232863}
H.~Li, D.~Serrano, T.~Guillod, S.~Wang, E.~Dogariu, A.~Nadler, M.~Luo, V.~Bansal, N.~K. Jha, Y.~Chen, C.~R. Sullivan, and M.~Chen, ``How magnet: Machine learning framework for modeling power magnetic material characteristics,'' \emph{IEEE Transactions on Power Electronics}, vol.~38, no.~12, pp. 15\,829--15\,853, 2023.

\bibitem{10535744}
Y.~Hu, J.~Xu, J.~Wang, and W.~Xu, ``Physics-inspired multimodal feature fusion cascaded networks for data-driven magnetic core loss modeling,'' \emph{IEEE Transactions on Power Electronics}, vol.~39, no.~9, pp. 11\,356--11\,367, 2024.

\bibitem{9773372}
H.~Li, D.~Serrano, T.~Guillod, E.~Dogariu, A.~Nadler, S.~Wang, M.~Luo, V.~Bansal, Y.~Chen, C.~R. Sullivan, and M.~Chen, ``Magnet: An open-source database for data-driven magnetic core loss modeling,'' in \emph{2022 IEEE Applied Power Electronics Conference and Exposition (APEC)}, 2022, pp. 588--595.

\bibitem{breiman2001random}
L.~Breiman, ``Random forests,'' \emph{Machine learning}, vol.~45, no.~1, pp. 5--32, 2001.

\bibitem{10.1145/2939672.2939785}
\BIBentryALTinterwordspacing
T.~Chen and C.~Guestrin, ``Xgboost: A scalable tree boosting system,'' in \emph{Proceedings of the 22nd ACM SIGKDD International Conference on Knowledge Discovery and Data Mining}, ser. KDD '16.\hskip 1em plus 0.5em minus 0.4em\relax New York, NY, USA: Association for Computing Machinery, 2016, p. 785–794. [Online]. Available: \url{https://doi.org/10.1145/2939672.2939785}
\BIBentrySTDinterwordspacing

\end{thebibliography}

\end{document}